\documentclass[runningheads]{llncs}
\usepackage[T1]{fontenc}
\usepackage{graphicx}
\usepackage{algpseudocode}
\usepackage{algorithm}
\usepackage[toc,page]{appendix}
\usepackage{float}
\usepackage{adjustbox}
\graphicspath{ {./figures/} }

\begin{document}
\title{Refutation of Spectral Graph Theory Conjectures with Search Algorithms}
%
%
\author{Milo Roucairol\textsuperscript{a} and Tristan Cazenave\textsuperscript{a}}
\authorrunning{M. Roucairol et al.}
%
\institute{\textsuperscript{a}LAMSADE, Université Paris-Dauphine, Paris, Pl. du Maréchal de Lattre de Tassigny, 75016 Paris, France}
\maketitle              
\begin{abstract}
We are interested in the automatic refutation of spectral graph theory conjectures. Most existing works address this problem either with the exhaustive generation of graphs with a limited size or with deep reinforcement learning. Exhaustive generation is limited by the size of the generated graphs and deep reinforcement learning takes hours or days to refute a conjecture. We propose to use search algorithms to address these shortcomings to find potentially large counter-examples to spectral graph theory conjectures in seconds. We apply a wide range of search algorithms to a selection of conjectures from Graffiti. Out of 13 already refuted conjectures from Graffiti, our algorithms are able to refute 12 in seconds. We also refute conjecture 197 from Graffiti which was open until now.

\end{abstract}

\section{Introduction}

\label{sec:intro}

Monte Carlo search algorithms have proven to be powerful as game-playing agents, with recent successes like AlphaGo \cite{Silver2016MasteringTG}. These algorithms have the advantage of only needing an evaluation function for the final state of the space they explore. 

Graph conjectures are propositions on graph classes (any graph, trees, $K_n-free$...) that are suspected to be true and are awaiting proof or a refutation. They lend themselves well to computer-assisted proofs, as finding a counter-example can be tedious to do manually. Spectral graph conjectures are appropriate for automated refutation because the property can often directly be turned into the evaluation function that can take many different values.
Thanks to software like Auto-GraphiX \cite{hansen_autographix_2000} and Graffiti \cite{delavina_history_nodate}, there are plenty of such conjectures.

Adam Zsolt Wagner showed that one could find explicit counter-examples using deep reinforcement learning of a policy with the deep cross entropy method \cite{wagner_constructions_2021}. Wagner was able to refute three conjectures from Graffiti. Wagner's method trained a neural network with reinforcement learning for each conjecture, learning to build graphs tailored to the refutation of each conjecture. The refutation of a conjecture with Wagner's method takes hours or days. Using search algorithms it was possible to refute conjectures 2.1 and 2.4 from the Wagner paper in one second and conjecture 2.3 in 291 seconds \cite{roucairol2022refutation}.

A usual way to find counter-examples to graph theory conjectures is to generate all the graphs smaller than a given size and to verify the conjecture for all these small graphs \cite{aouchiche_survey_2010}. This exhaustive generation limits the size of the graphs that can be tested. On the contrary, our search algorithms can rapidly generate larger refutation graphs.

In this paper, the search algorithms will play the game of refuting conjectures by building counter-examples edge by edge. We build on our previous paper \cite{roucairol2022refutation} by testing many more conjectures and by comparing more search algorithms.

First, we will present the refutation of graph theory conjectures, then the different algorithms we use to explore the problem space, after that the procedure we use to build graphs and the game rules, and finally we expose our new results on multiple different conjectures.

\section{Refutation of Graph Theory Conjectures}
\label{sec:refutation}

\subsection{Graph Theory Conjectures}
\label{sec:GTC}

Refuting Graph theory conjectures can be a hard task to do manually. Imagining a large number of graphs and computing invariant or NP-hard problem values can be tedious and often results in a waste of time. Computers are designed to help with these score computations. The goal of automated conjecture refutation is to automatize the exploration as well.\\

Graph theory conjecture can concern many different properties of graphs: existence, topology, flow, connectivity, cycle, minors, spectral...
Here we are interested in the spectral graph theory conjectures, necessitating matrix calculations only. The spectrum of a matrix is the set of its eigenvalues. Spectrum-related invariants on different types of graph-related matrices (adjacency, distance, Laplacian...) are the focus of spectral graph theory conjectures.\\

We propose to compare search algorithms for the refutation of various spectral graph theory conjectures. In \cite{roucairol2022refutation}, the conjectures from \cite{wagner_constructions_2021} were refuted hundreds of times faster and the performances of the methods were comapred. This paper experiments with a more diverse set of search algorithms and we address all the remaining conjectures from Graffiti that do not involve solving a NP-hard problem. Graffiti is a well-known software that was used to generate a thousand spectral graph theory conjectures. It was the target of numerous works and discussions, sometimes involving some of the greatest graph theoricians such as Paul Erdos, Laszlo Lovasz, Noga Alon, Noam Nisan. Conjectures from Graffiti were the subject of entire articles, with graph theorists dedicating weeks or months to them. Our program aims at refuting some of them in a much more efficient manner.

We decided to focus on this subset of Graffiti conjectures to avoid dependence on other algorithms and to reduce the number of conjectures.

\subsection{Algorithms Used to Refute Graph Conjectures}
\label{sec:AURGC}

The program by Wagner trained a deep neural network with cross-entropy to output a policy. The network is used to learn a policy from a state. The policy is used the generate a batch of graphs, it then evaluates the best graphs from the batch and trains the neural network using the graph's scores. Batches are successively evaluated until the refutation of the conjecture.\\

Softwares used for automated conjecture generation like Graffiti or Auto-GraphiX also have their own ways of automatically refuting conjectures to verify the easily refutable ones.
Graffiti and its conjectures are detailed in Aouchiche and Hansen survey \cite{aouchiche_survey_2010}. It generates many conjectures in the form of inequalities between invariants. It then tests these conjectures on a database of graphs and discards the falsified ones. Then it checks if the inequalities are not implied by already known theorems and conjectures. If a conjecture passes these tests, it is proposed to scientists. "Written on the wall" collects almost 1000 conjectures from Graffiti, along with the discussions of many scientists. Graffiti's conjectures connect various topics in spectral graph theory such as eigenvalues, adjacency, distance; Laplacian, and gravity matrices of graphs. These conjectures are the subjects of dozens of articles.


\section{Materials and Methods}

\label{sec:newalgoconj}
We decided to expand the experiments from \cite{roucairol2022refutation} by using the same algorithms and adding two widely used Monte Carlo Search algorithms, another recent one, and another baseline simple algorithm on more conjectures produced by Graffiti \cite{aouchiche_survey_2010}. The search model used by the algorithms to build the graphs is the same as in \cite{roucairol2022refutation}: starting from a single node, new nodes and edges are added to the graph until it reaches a target size. These additions of nodes and edges are the edges of the search tree.

\begin{itemize}
    \item NMCS \cite{CazenaveIJCAI09} uses nested levels of search with random search at the base level. It is already used in \cite{roucairol2022refutation}.
    \item NRPA \cite{rosin_nested_2011} learns a playouts policy with nested levels of best sequences. At the lowest level it makes playouts with the learned policy. It is already used in \cite{roucairol2022refutation}.
    \item GBFS is a simple greedy algorithm opening the best state from a list, evaluating the children of this state, and inserting these children in the list according to their evaluation. It is already used in \cite{roucairol2022refutation}.
    \item BEAM search, another baseline greedy search heuristic, keeps the $width$ best states after each step of expanding and going down the search tree.
    \item Upper Confidence bounds applied to Trees (UCT) as described by Kocsis and Szepesvári \cite{Kocsis2006}, which is the most widely used MCTS algorithm, and close to PUCT used in outstanding works such as Deepmind's Alphazero \cite{silver2017mastering}.
    \item Rapid action value estimation (RAVE), a MCTS algorithm inspired by UCT and proposed by Siver and Gelly \cite{gelly2011monte}, then generalized as GRAVE by Cazenave \cite{cazenave2015generalized} using a threshold on the number of playouts.
    \item Lazy Nested Monte Carlo Search (LNMCS), a variant of NMCS proposed by Roucairol and Cazenave \cite{roucairol2023solving} addresses a shortcoming of the NMCS. Before launching a costly lower level LNMCS on a child state, it is evaluated with playouts and pruned if the evaluation is not satisfying enough, this way it avoids sinking the time budget in fruitless subtrees and allows the LNMCS to use higher starting levels than the NMCS.
\end{itemize}

These conjectures (in table \ref{tableNew}) were selected because they did not require NP-hard problems to be solved to compute the score, thus lending themselves well to Monte Carlo Search and other optimization methods. All of the fitting conjectures from Aouchiche and Hansen's survey \cite{aouchiche_survey_2010} are included in our experiments. It is possible to tackle conjectures needing a NP-hard problem to be solved (like coloration) up to sizes 25-30, we might come back to such conjectures in future works.

\section{Results}
\label{sec:newresults}

\begin{table}[h]
\caption{Times in seconds to obtain a refutation by applying every algorithm on selected Graffiti conjectures.}
\label{tableNew}
\begin{center}
\begin{adjustbox}{angle=90}
\begin{tabular}{l | l | l | l l l l l l l l } 
\hline\noalign{\smallskip}
 Graffiti & Size & Graph type & NMCS & LNMCS & NRPA & UCT & GBFS & BEAM & GRAVE & RAVE\\ 
 \noalign{\smallskip}\hline\noalign{\smallskip}
 166 R & 5+ & any & 0 & 0 & 0 & 0 & 0 & 0 & 0 & 0  \\

 321 R & 4+ & $K3-free$ & 0 & 0 & 0 & 0 & 0 & 0 & 0 & 0  \\

 189 R & 5+ & any & 0 & 0 & 0 & 2 & 10 & 4 & 2 & - \\

 289 R & 20+ & girth $\geq 5$ & 600 & 600 & 200 & - & 6 & 102 & - & -  \\

 302 R & 6+ & tree & 0 & 0 & 0 & 0 & 0 & 0 & 0 & 0 \\

 715 R & 16+ & tree & 0 & 0 & 0 & 0 & 1 & 0 & 0 & 0 \\

 29 R & 7+ & any &  2 & 2 & 10 & 5 & 0 & - & 2 & 1 \\

 30 R & 12+ & any &  0 & 0 & 0 & 0 & 311 & - & 0 &  0 \\

 301 R & 14+ & tree &  2 & 2 & 4 & - & 0 & 0 & 7 &  2 \\

 137 R & 67+ & any &  - & - & - & - & 513 & - & - &  - \\

 139 R & 50+ & any & - & - & - & - & 36 & - & - & - \\

 711 R & 5+ & any & 0 & 0 & 4 & 0 & 0 & 0 & 0 & 0 \\
 
197 O & 17+ & $K3-free$  & 30 & 30 & 5 & - & 0 & 4 & - & - \\

 140 R & 50 & any \& tree & - & - & - & - & - & - & - & -  \\

 290 O & 50 & girth $\geq 5$ & - & - & - & - & - & - & - & - \\

 284 O & 50 & girth $\geq 5$ & - & - & - & - & - & - & - & - \\

 195 O & 50 & any & - & - & - & - & - & - & - & - \\

 21 O & 50 & any \& tree & - & - & - & - & - & - & - & - \\

 39 O & 50 & any \& tree & - & - & - & - & - & - & - & - \\

 143 O & 100 & any \& tree & - & - & - & - & - & - & - & - \\

 154 O & 50 & any \& tree & - & - & - & - & - & - & - & - \\

 20 O & 50 & any \& tree & - & - & - & - & - & - & - & - \\

 40 O & 50 & any \& tree & - & - & - & - & - & - & - & - \\

 254 O & 50 & any \& tree & - & - & - & - & - & - & - & - \\

 262 O & 50 & any \& tree & - & - & - & - & - & - & - & - \\

 712 O & 50 & any \& tree & - & - & - & - & - & - & - & - \\

 198 O & 50 & any  & - & - & - & - & - & - & - & - \\

 714 O & 100 & any  & - & - & - & - & - & - & - & - \\

 219 O & 50 & $K3-free$ & - & - & - & - & - & - & - & -  \\

 322 O & 50 & $K3-free$ & - & - & - & - & - & - & - & -  \\

 292 O & 50 & girth $\geq 5$ & - & - & - & - & - & - & - & -  \\

295 O & 50 & girth $\geq 5$ & - & - & - & - & - & - & - & -  \\

129 O & 50 & any  & - & - & - & - & - & - & - & -  \\

698 O & 50 & any  & - & - & - & - & - & - & - & -  \\

252 O & 50 & any  & - & - & - & - & - & - & - & -  \\

\noalign{\smallskip}\hline
\end{tabular}
\end{adjustbox}
\end{center}

Graffiti:
R means the conjecture was already refuted (as reported in \cite{aouchiche_survey_2010})
O means the conjecture is open to be proved or refuted

Size:
If there is a "+", it is the size at which we start to find counter-examples with our algorithms. Otherwise, it is the size we built graphs up to.

Graph type:
It is the types of graphs we used in our searches, "any" means there was no restriction on the graph built, $K3-free$ means there is no clique of size 3, "girth $\geq$ 5" means the shortest cycle must be longer than 5, "tree" means the graph must be a tree. We tried both generating trees and then any graph on multiple conjectures.

A 0 in the time section means that the conjecture was refuted in less than a second.

\end{table}

The experiments were made with Rust 1.59, on an Intel Core i5-6600K 3.50GHz using a single core (but parallel processing is very accessible).

Our positive results in table \ref{tableNew} were obtained by launching each algorithm on each conjecture with a time budget of 15 minutes. In the Graffiti column, R means the conjecture was already refuted, and O means the conjecture is still open. 
\begin{itemize}
    \item The NMCS used a level of 3.
    \item The LNMCS used a level of 4, 3 playouts per evaluation, and a ratio of 0.8.
    \item The UCT used a constant of 1.
    \item The GRAVE and RAVE used a reference of 5.
    \item The BEAM search used a width of 10.
\end{itemize}

\section{Discussion}

As shown in table \ref{tableNew}, we were unable to refute conjectures that were open except one (either 197 or 322). This means either that our algorithms are not strong enough, or that these conjectures are true. However, with the exception of Graffiti 140, most of the already refuted conjectures we tried were refuted by at least one algorithm among GRAVE, RAVE, NMCS, LNMCS, NRPA, GBFS, and UCT. Graffiti 140 refutation seems to be lost, 140 is said to be refuted by another work by Favaron which we are unable to find.

Considering nonterminal states for evaluation helped immensely with the harder Graffiti 137, 139, 189, 289, and 301. Graffiti 289's repeating pattern may be an explanation as to why the NRPA succeeded where most other algorithms failed. The times, when displayed in the hundreds of seconds, are approximations, only GBFS and BEAM are deterministic so the time is exact. In addition, NMCS, UCT, GRAVE, and RAVE were very unreliable at finding a counter-example for Graffiti 29, about once every 3 runs, when LNMCS was able to find a counter-example in less than 60 seconds in 9 out of 10 runs. 

For Graffiti 189 and 195, we first searched in the "any graph" search space (instead of the type of graphs the conjecture is about), and then verified if the graph in question complied with the conjecture definition.

We think the relative failure of GBFS on Graffiti 30 is partly due to the score function leading the algorithm to open nodes to smaller graphs first, while the MCTS algorithms try larger graphs immediately using playouts. Beam search requires a width of 80 or more to find a counter example to this conjecture, so trying larger graphs is not sufficient on its own.

The beam search is very dependent on its width parameter, too high and it will not be able to explore deep in the search tree and find large counter examples (on Graffiti 137), and too low and it will not consider inferior intermediate states that would in fact lead to a counter example (on Graffiti 29 and 30). That makes it an inferior choice for spectral graph conjecture refutation compared to GBFS.

\begin{figure}[h]
    \centering
    \includegraphics[width=12cm]{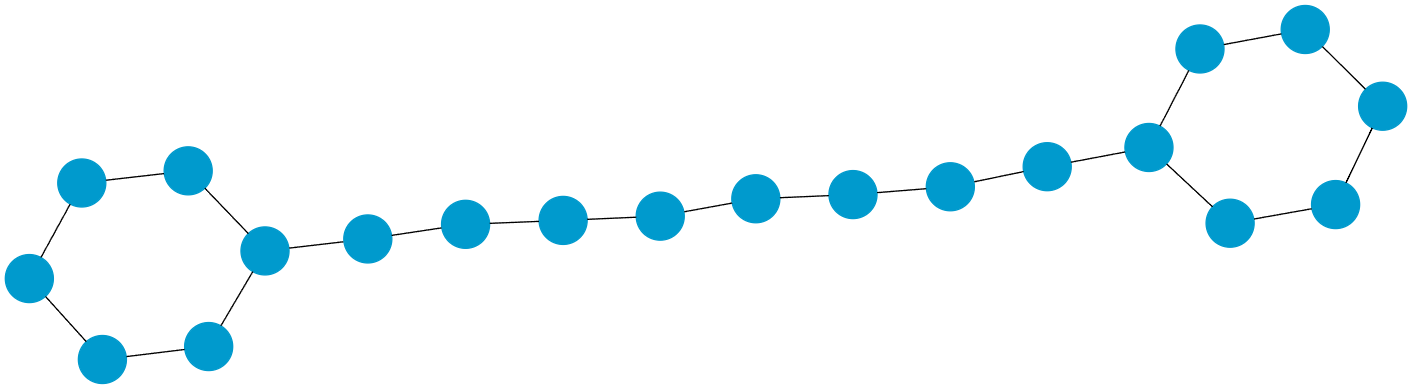}
    \caption{A counter-example of Graffiti 289 of size 20 (second largest eigenvalue $\le$ mean of the mean of all adjacent vertex degree for all nodes)}
    \label{graph289}
    \raggedright
        Edges: 0-1, 1-2, 2-3, 3-4, 4-5, 5-6, 6-7, 7-8, 8-9, 9-10, 10-5, 0-11, 11-12, 12-13, 13-14, 14-15, 15-16, 16-17, 17-18, 18-19, 19-14

\end{figure}

\begin{figure}[H]
    \centering
    \includegraphics[width=5cm]{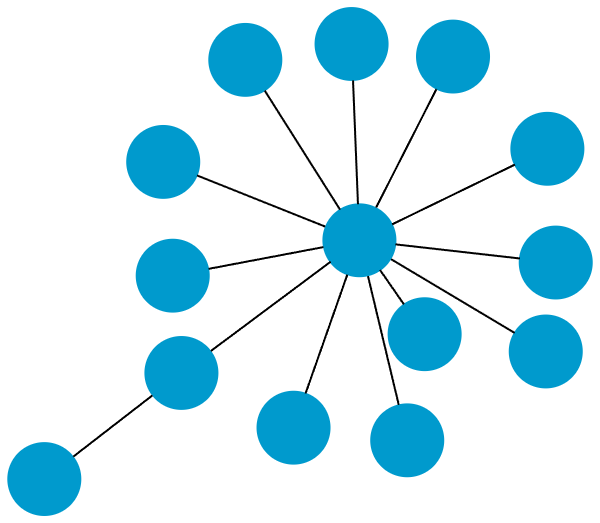}
    \caption{A counter-example of Graffiti 301 of size 14 (scope of positive eigenvalues $\le$ harmonic)}
    \label{graph301}
    \raggedright
    Edges: 0-1, 0-2, 1-3, 1-4, 1-5, 1-6, 1-7, 1-8, 1-9, 1-10, 1-11, 1-12, 1-13

\end{figure}

\begin{figure}[H]
    \centering
    \includegraphics[width=5cm]{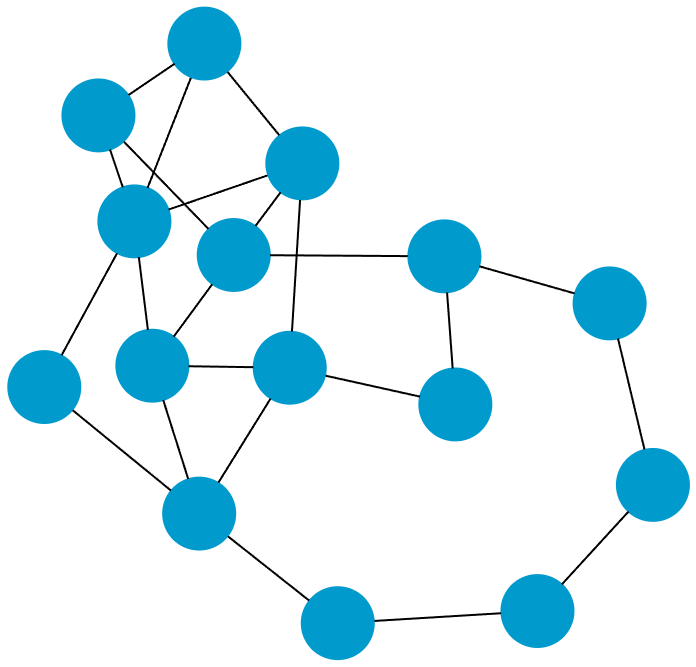}
    \caption{A counter-example of Graffiti 30 of size 15 (\# positive distance eigenvalues $\le$ sum of temperatures)}
    \label{graph30}
    \raggedright
    Edges: 0-1, 0-3, 0-12, 1-2, 1-3, 2-3, 2-7, 2-12, 3-4, 3-5, 4-6, 5-6, 5-7, 5-12, 6-7, 6-8, 7-9, 8-10, 9-11, 10-14, 11-12, 11-13, 13-14

\end{figure}

\begin{figure}[H]
    \centering
    \includegraphics[width=3cm]{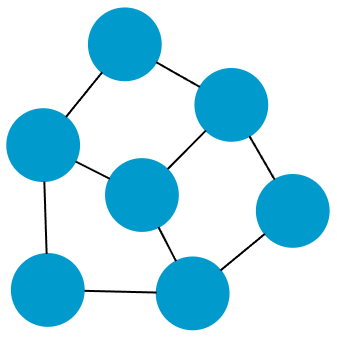}
    \caption{A counter-example of Graffiti 29 of size 7 (randic index $\le$ \# negative eigenvalues)}
    \label{graph29}
    \raggedright
Edges: 0-1, 0-2, 0-6, 1-3, 2-5, 3-4, 3-6, 4-5, 5-6
\end{figure}

\begin{figure}[p]
    \centering
    \includegraphics[width=8cm]{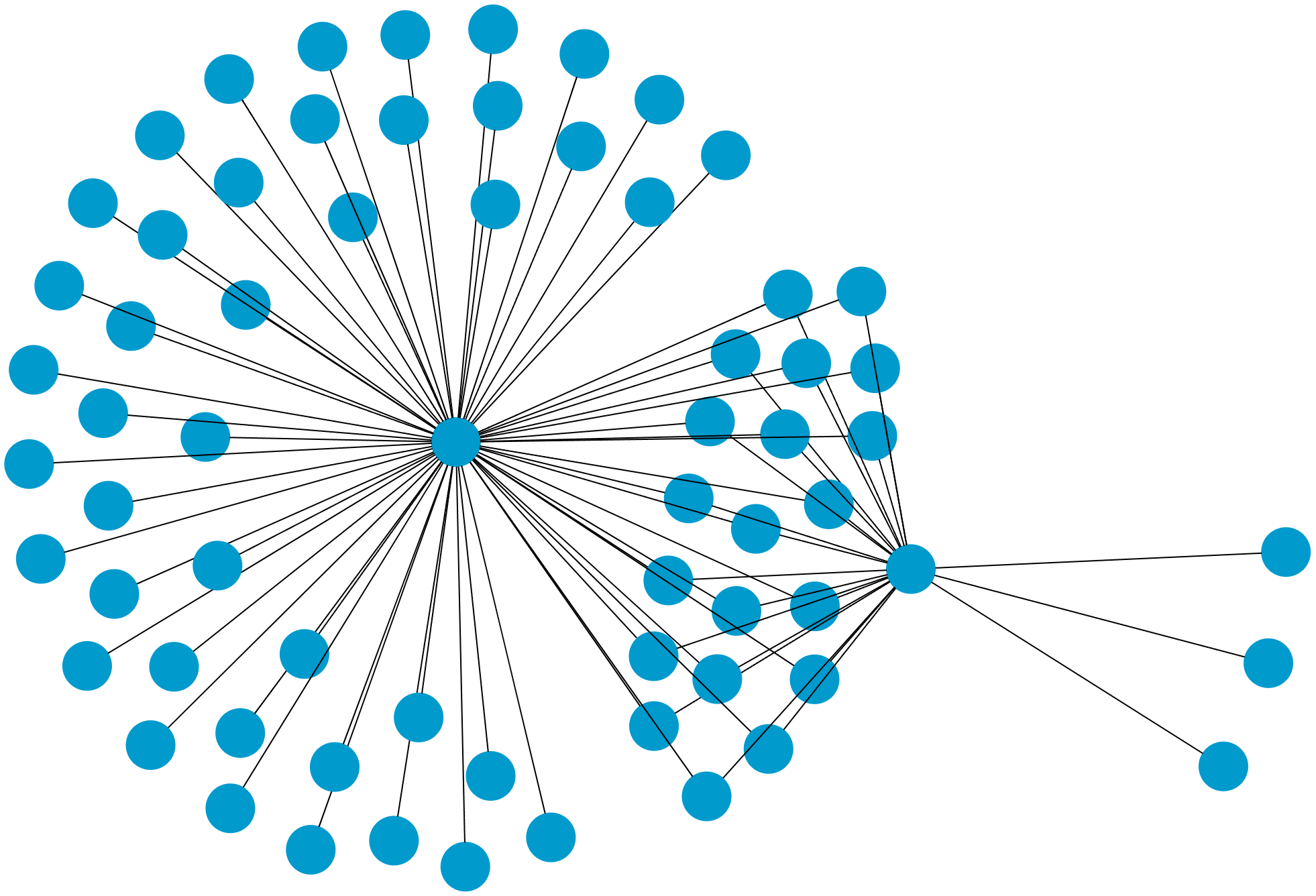}
    \caption{A counter-example of Graffiti 137 of size 67 (second largest eigenvalue $\le$ harmonic)}
    \label{graph137}
    \raggedright

Edges: 0-1, 0-2, 1-3, 1-4, 1-5, 1-11, 1-15, 1-18, 1-21, 1-25, 1-28, 1-31, 1-34, 1-37, 1-40, 1-43, 1-46, 1-49, 1-52, 1-55, 1-58, 1-61, 1-64, 1-66, 2-6, 2-7, 2-8, 2-9, 2-10, 2-11, 2-12, 2-13, 2-14, 2-15, 2-16, 2-17, 2-18, 2-19, 2-20, 2-21, 2-22, 2-23, 2-24, 2-25, 2-26, 2-27, 2-28, 2-29, 2-30, 2-31, 2-32, 2-33, 2-34, 2-35, 2-36, 2-37, 2-38, 2-39, 2-40, 2-41, 2-42, 2-43, 2-44, 2-45, 2-46, 2-47, 2-48, 2-49, 2-50, 2-51, 2-52, 2-53, 2-54, 2-55, 2-56, 2-57, 2-58, 2-59, 2-60, 2-61, 2-62, 2-63, 2-64, 2-65, 2-66\\
\end{figure}

In\cite{roucairol2022refutation} Graffiti 137 was one of the 4 main conjectures refuted. Unfortunately an erroneous definition of the Harmonic was used (see subsection \ref{sec:erratum}). Graffiti 137 is actually more interesting than we initially thought, the smallest known counter-example featured 101 nodes and required a very specific structure, it was first refuted by Favaron et al. \cite{FAVARON1993197} using a mathematical proof and not directly a counter-example. Here we managed to find a counter-example of size 67 (see figure \ref{graph137}) which does not follow the structure described, but closely resembles it, by descending the search space (opening graphs of increasing size, much like conjecture 2) almost directly with GBFS.

\begin{figure}[p]
    \centering
    \includegraphics[width=10cm]{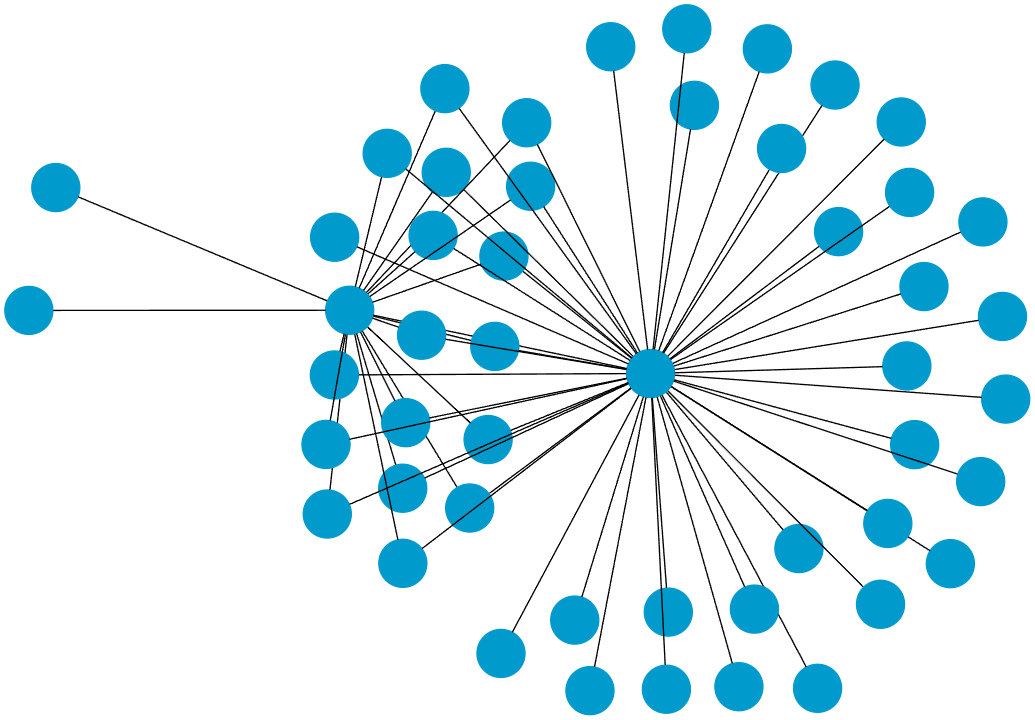}
    \caption{A counter-example of Graffiti 139 of size 50 ( - second smallest eigenvalue $\le$ harmonic)}
    \label{graph139}
    \raggedright

Edges: 0-1, 0-2, 0-3, 0-4, 0-5, 0-8, 0-11, 0-14, 0-17, 0-20, 0-23, 0-26, 0-29, 0-32, 0-35, 0-38, 0-41, 0-44, 0-47, 0-49, 1-2, 2-3, 2-6, 2-7, 2-8, 2-9, 2-10, 2-11, 2-12, 2-13, 2-14, 2-15, 2-16, 2-17, 2-18, 2-19, 2-20, 2-21, 2-22, 2-23, 2-24, 2-25, 2-26, 2-27, 2-28, 2-29, 2-30, 2-31, 2-32, 2-33, 2-34, 2-35, 2-36, 2-37, 2-38, 2-39, 2-40, 2-41, 2-42, 2-43, 2-44, 2-45, 2-46, 2-47, 2-48, 2-49

\end{figure}

As mathematical proofs are complex, if the goal was to refute this one conjecture, then our algorithm was most likely faster. But another interesting result is that the class of graph needed to refute Graffiti 137 by Favaron et al. was also used to refute two other conjectures (and could have been used to refute Graffiti 139, see figure \ref{graph139}), the search algorithms can provide insight to the mathematicians (they can provide insight even if the graph produced does not refute a conjecture as in Wagner's work \cite{wagner_constructions_2021} with conjecture 2.3).
The other algorithms did not manage to find a counter-example in a reasonable time.

\subsection{Graffiti 197 is refuted}

In "Written on the wall", Graffiti 197 is defined as: \\

\textbf{Graffiti 197.}  - 2-nd smallest eigenvalue $\leq$ range of eigenvalues of gravity matrix. \\

\begin{figure}[h]
    \centering
    \includegraphics[width=5cm]{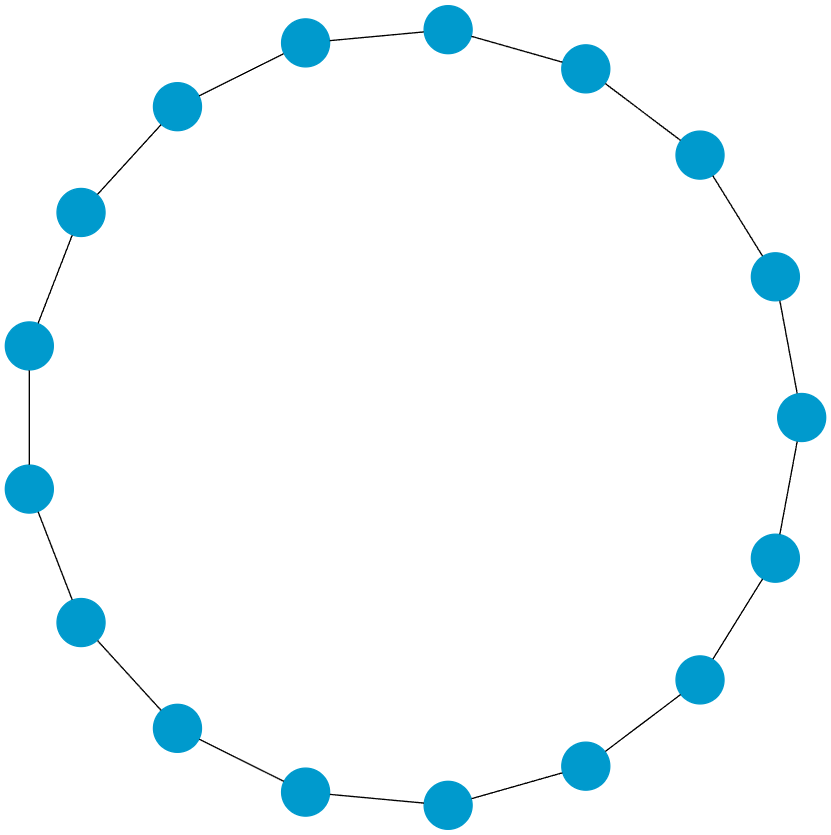}
    \caption{A counter-example of Graffiti 197 of size 17}
    \label{graph197}
    \raggedright

Edges: 0-1, 1-2, 02-3, 3-4, 4-5, 5-6, 6-7, 7-8, 8-9, 9-10, 10-11, 11-12, 12-13, 13-14, 14-15, 15-16, 16-0

\end{figure}

Using the usual definition of the range, and not Aouchiche and Hansen's (see subsection \ref{sec:erratum}), our algorithms find a counter example with the cycle of length 17, see figure \ref{graph197}. Using Aouchiche and Hansen's definition of the range we could not find any counter example, but if this initial definition was to be used then Graffiti 322 would be (too) easily refuted.

The second smallest eigenvalue is $\lambda_{n-1} \approx -1.9659 $, and the range of eigenvalues of the gravity matrix is approximately 1.7035.

Cycles of size 21 and 25 also refute Graffiti 197, we conjecture that any cycle of size $1 + 4*n$ greater or equal to 17 refutes this conjecture.

\subsection{Erratum} \label{sec:erratum}

We provide a copy of "Written on the Wall" on this project's github, which is otherwise hard to find: \\

\begin{itemize}
    \item Graffiti 290 (open) was solved instantly using the definition of gravity in Aouchiche and Hansen's survey \cite{aouchiche_survey_2010} (which prompted us to verify the definition). But it was seemingly impossible using the definition from Brewster et al. \cite{brewster1995computational}.

Since the definition of gravity of a graph is not widespread, here is the correct definition of the gravity matrix from "Written on the Wall" (page 52) and Brewster et al. \cite{brewster1995computational}: \\

\textbf{Gravity matrix:} The matrix (indexed by vertices of the graph) whose $(u, v)th$ entry is 0 if $u=v$ or if there is no path joining u to v; otherwise it is

\begin{equation}
Gr(G) = (1/(n - 1))(d(u) * d(v))/d(u, v)
\end{equation}

Where $d(u)$ is the degree of vertex $u$ and $d(u, v)$ is the distance from vertex $u$ to vertex $v$ in the graph $G$ of size $n$.

\item In the same manner, with Aouchiche and Hansen's definition of the harmonic, Graffiti 140 was refuted by the graph with one edge connecting two vertices. The correct definition is featured in "Written on the Wall" (page 28) and in Favaron et al. \cite{FAVARON1993197}: \\

\textbf{Harmonic:} The harmonic of a graph is defined as:
\begin{equation}
Hc(G) = \sum_{uv \in E} \frac{2}{d(u) + d(v)}
\end{equation}

Where and $E$ is the set of edges of a graph $G$.\\

\item Using Aouchiche and Hansen survey's \cite{aouchiche_survey_2010} definition of range, it appeared that our algorithms managed to refute Graffiti 322 (open) with a cycle graph of size 4. Here is the definition of Graffiti 322 from "written on the wall": \\

\textbf{Graffiti 322.} If G is a triangle-free graph then the Inverse Even $\leq$ range of eigenvalues of Distance. \\

Inverse Even is defined  as the sum over the vertices of the inverse of the number of vertices to even distance from that vertex, ie:

$\sum_{v \in V} 1/Ev(v) $ where $Ev(v)$ is the number of vertices at even distance from vertex $v$.

Thus the cycle of size 4 has 3 distinct distance eigenvalues and an Inverse Even of 4. An error with the definitions seems more likely than this conjecture being left open after dozens of articles with such a simple counter-example.

The range is defined as the number of distinct values in a vector in Aouchiche and Hansen's survey and (likely inherited from) Favaron et al.'s work, with no further sources. The range of a vector is usually the difference between the minimum and maximum values. Thus we think this definition is erroneous but we were not able to find the actual definition of the range used in "written on the wall". The results featured in table \ref{tableNew} for Graffiti 322 use the usual definition of range, either way, Graffiti 197 is refuted or Graffiti 322 is refuted.

\item Graffiti 294 from Aouchiche and Hansen's survey is Graffiti 295 from "written on the wall", we go by this identifier in table \ref{tableNew}.

\item Graffiti 140 is said to be refuted by Favaron et al. in "residue of a graph" \cite{favaron1991residue} according to "written on the wall", but there is no mention of "harmonic", "deviation" or Graffiti 140 in that paper.

\end{itemize}


The general lack of sourcing, the loss of multiple articles, and the misleading or erroneous definitions are detrimental to future work on Graffiti conjectures. For instance, we suspect that Graffiti conjectures 28 and 209 may not be refuted as sources are nonexistent (28) or seemingly lost (209). We decided to include Graffiti 140 in the table unlike 28 and 209 because we could find its refutation source, but we could not find its appearance in the source. 
Auto-GraphiX conjectures may be more favorable to continuing this work.

\section{Conclusion}\label{sec:conclusion}

Search algorithms proved to be powerful ways of refuting conjecture from spectral graph theory, much faster than Wagner's deep cross entropy method \cite{wagner_constructions_2021}. We can identify three algorithms that seem more effective on the conjectures tested: GBFS which is overall faster but was not able to refute conjectures in \ref{roucairol2022refutation} and struggled on Graffiti 30,  NRPA which is very efficient when a repeating pattern refutes a conjecture, and LNMCS/NMCS which managed to be slightly faster than NRPA on Graffiti 29 and 301 while being faster than the other MCTS algorithms. LNMCS did not outperform NMCS due to the nature of the search space: there rarely is a move that doom all its child-states It seems that not one single algorithm is able to solve all of sample of conjectures we selected, and using different approaches is required to adapt to the variety of problems in spectral graph theory conjecture refutation. We recommend trying GBFS, the NMCS, and then NRPA.

Trying to build trees even when the conjecture is applied on any graph can also be helpful as it reduces the amount of possible builds greatly and focuses on a sub-type of graphs that often features extreme properties over the spectral invariants. It is inexpensive and should be tried first.

However, these methods present limits. Computing score functions that require eigenvalues on big trees (over size 300) can be very costly. They are also dependent on the shape of the score function: a noisy score function with many local minimums can be challenging, as well as a score function with more discrete results can lead to an absence of differentiation in the paths to explore. Conjectures requiring computing a NP-hard problem can also severely increase the computing time even for small graphs (30 vertices).

Despite these limitations, and depending on the definition of the gravity matrix, either our program was able to refute up to the state of the art (manually from graph theorists) the conjectures from Graffiti, or it refuted Graffiti 197, a previously open conjecture.

You can access the implementations of the conjecture and the Rust code used to refute them here: \\

https://github.com/RoucairolMilo/refutationExperimentalMathematics

\section*{Statements and Declarations}

Thanks to \'Edouard Bonnet for fruitful discussions. Thanks to Craig E. Larson and Ermelinda DeLaVi\~na for answering some of our questions and providing "written on the wall" to us. 

\subsection{funding} This work was supported in part by the French government under the management of Agence Nationale de la Recherche as part of the “Investissements d’avenir” program, reference ANR19-P3IA-0001 (PRAIRIE 3IA Institute).

\subsection{competing interests}
The authors have no relevant financial or non-financial interests to disclose.

\bibliographystyle{plain}
\bibliography{refs}

\end{document}